\documentclass[conference]{IEEEtran}
\IEEEoverridecommandlockouts
\usepackage{cite}
\usepackage{amsmath,amssymb,amsfonts}
\usepackage{algorithmic}
\usepackage{graphicx}
\usepackage{textcomp}
\usepackage{xcolor}
\usepackage{flushend} 
\usepackage{hyperref} 
\def\BibTeX{{\rm B\kern-.05em{\sc i\kern-.025em b}\kern-.08em
    T\kern-.1667em\lower.7ex\hbox{E}\kern-.125emX}}
    
\begin{document}

\title{3D Convolution on RGB-D Point Clouds for Accurate Model-free Object Pose Estimation \\
\thanks{The authors are with the School of Mechanical and Aerospace Engineering, Nanyang Technological University, Singapore.
}
}

\author{\IEEEauthorblockN{Zhongang Cai}
\and
\IEEEauthorblockN{Cunjun Yu}
\and
\IEEEauthorblockN{Quang-Cuong Pham}
}

\maketitle

\begin{abstract}
The conventional pose estimation of a 3D object usually requires the knowledge of the 3D model of the object. Even with the recent development in convolutional neural networks (CNNs), a 3D model is often necessary in the final estimation. In this paper, we propose a  two-stage pipeline that takes in raw colored point cloud data and estimates an object’s translation and rotation by running 3D convolutions on voxels. The pipeline is simple yet highly accurate: translation error is reduced to the voxel resolution (around 1 cm) and rotation error is around 5 degrees. The pipeline is also put to actual robotic grasping tests where it achieves above 90\% success rate for test objects. Another innovation is that a motion capture system is used to automatically label the point cloud samples which makes it possible to rapidly collect a large amount of highly accurate real data for training the neural networks. 
\end{abstract}

\section{Introduction}
Knowing the position and orientation of a target object is important for subsequent robotic manipulation tasks. The conventional approach compares the known model of the object with the image or point cloud captured to determine its pose. However, an accurate CAD model is often not readily available \cite{saxena_2008_ijrr}. Classical algorithms that work on point clouds such as principal component analysis (PCA) and iterative closest points (ICP) are prone to sensor noises and occlusion. ICP takes exponentially longer time with an increase in the number of points to match and it easily gets stuck in a local optimum if the initial guess is poor. 

Neural networks are exceptional in interpreting complex features amidst noises. Hence, we propose a pipeline takes in raw XYZRGB point clouds captured by a RGB-D sensor, and output directly the pose estimation including both translation and orientation. 3D convolution on point clouds for pose estimation has not been widely applied because of the difficulty of producing accurate labels and the high computation cost. To counter the former, we propose to use a motion capture (MoCap) system for automatic pose labelling; for the latter, we designed the pipeline to have two stages of estimation, allowing for larger and coarser voxels to be used for rough position estimation and smaller and finer voxels for rotation estimation.

In the first stage, a neural network estimates the translation of the object from the voxels generated from the entire point cloud that is bounded by the workspace; the second stage takes a closer look at the location where the object is estimated to be by the previous stage, crop out a sub-cloud, voxelize it with a finer grid size and pass it to a neural network for rotation estimation.

The main contributions of this paper are: 
\begin{itemize}
	\item we validate the proof-of-concept idea of direct pose estimation using 3D convolution on point clouds. The experiments have proved the effectiveness of this method to achieve high accuracy estimation.
	\item we present a 3D-2D neural network architecture that allows networks that are designed for 2D image applications to be directly used for 3D tasks.
	\item we design a two-stage estimation pipeline that requires simple neural networks and can run on a less powerful GPU, yet achieving very high accuracy.
	\item we propose a new pose labelling method using a MoCap system that makes collection and labelling of a large real 3D dataset possible. To our knowledge, this is the first attempt to fully utilizes a MoCap's flexibility in sample collection.
\end{itemize}


\begin{figure}[t]
 \centerline{\includegraphics[width=\columnwidth]{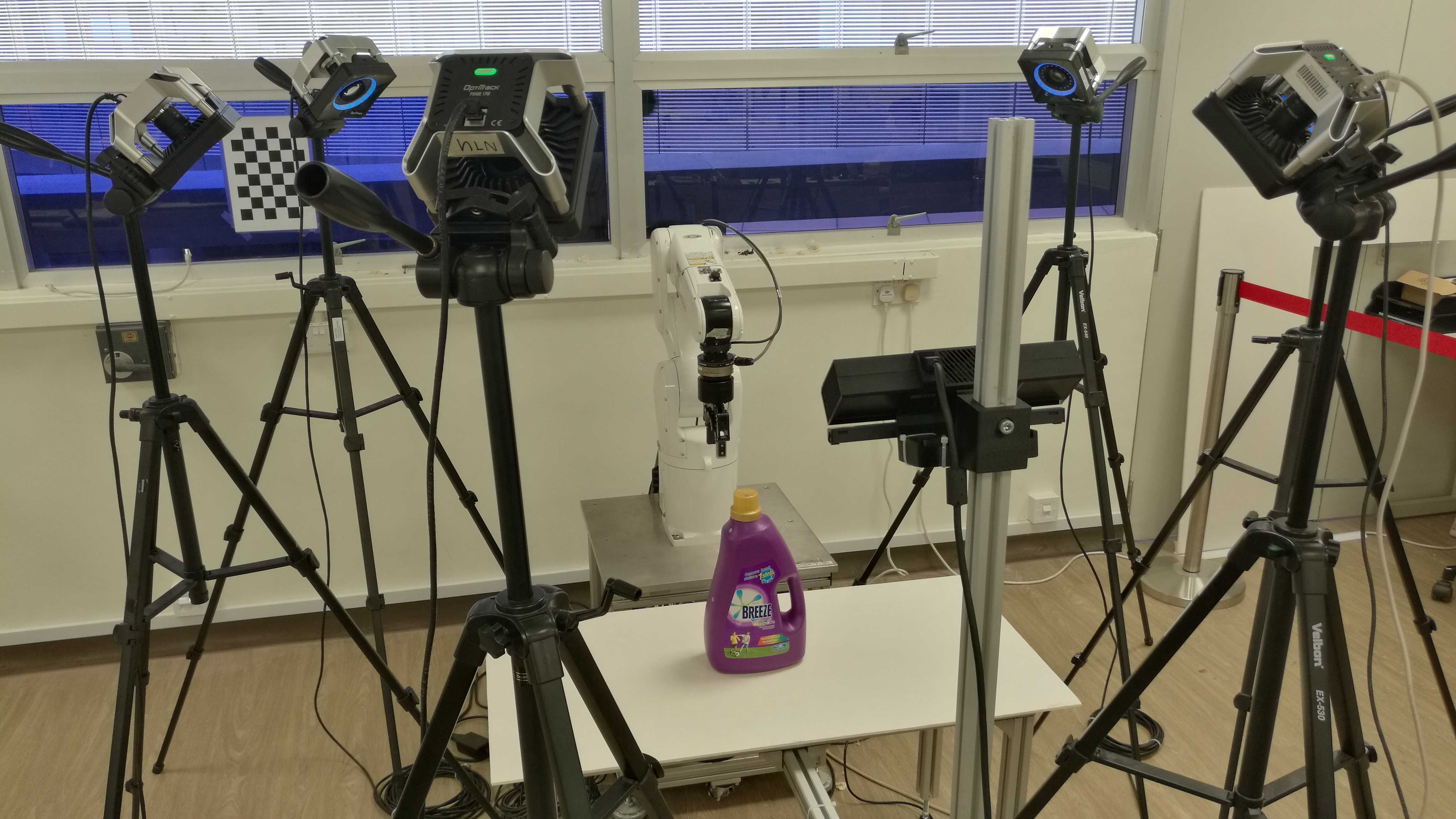}}
 \caption{Set-up: a test object is placed on the workbench between the RGB-D sensor and the robotic arm, and surrounded by MoCap cameras}
 \label{fig_setup}
\end{figure}

\section{Related Work}
\subsection{Classical Methods}
Classical methods towards pose estimation often require some prior knowledge of the target object (a CAD model or a scanned mesh model) and an iterative method has to be used in order to provide the final estimate. For example,  one can register the known 3D CAD model of the object on the captured image, optimizes the pose estimation through iteratively minimizing the error of misalignment \cite{jayawardena_2011_dicta}. Pose estimation using RGB-D sensors have also been explored prior to the prevalent use of deep learning \cite{choi_2012_iros}. However, these classical methods are limited by the availability of the models and achieve relatively low accuracy.


\subsection{Deep Learning on Images}
The advent of deep learning has opened up a new possibility to tackle the pose estimation problem. Some research effort has been put in pose estimation based on 2D image\cite{tremblay_2018_corl}, but bounding box coordinates instead of pose parameters are estimated. \cite{xiang_2018_rss} performs regression on quaternions, however, without refinement, the error is too large (less than 50\% angle estimation has less than 10 degrees) to be useful for robotic manipulation tasks such as grasping.

\subsection{Deep Learning on Point Clouds}
Researchers also take advantage of deep learning's robustness against noisy RGB-D point clouds and apply techniques in the semantic segmentation as part of the grasping pipeline \cite{mahler_2016_iros}\cite{zeng_2017_icra}\cite{wong_2017_iros}. However, an object model is required towards the end and classical techniques such as PCA and ICP are still used for the pose estimation of the object. \cite{xiang_2018_rss} also uses RGB-D data, but it is only for refinement and ICP is used.

With the advent of autonomous vehicles, for which LiDARs provide point clouds as an important data input, many have applied deep learning in the pose estimation of nearby vehicles. \cite{chen_2017_cvpr} We have taken inspirations from these works. With additional color information provided by RGB-D sensor and a much denser point cloud, the pose estimation of our network is expected to achieve better results.

\subsection{Deep Learning on Graspable Areas}
Specific to robotic grasping, deep learning has been applied to identify the graspable areas \cite{saxena_2008_ijrr}. However, it does not provide pose information of the object. In addition, when the graspable areas are not visible to the sensor (for example, when the handle is hidden behind the object), the approach does not work due to the fact that the neural network is not trained to understand the object.

\subsection{Other Deep Learning Methods}
There are also works to combine autoencoders and CNN\cite{inoue_2018_icip}. However, only translation estimation is addressed and manual labelling of real images are required.


\section{Set-up and Calibration}
The workbench is a $1 m\times0.5 m$ piece of white rigid board, fixed in front of the robot. Throughout the experiments, we are only interested in the object that appears on the workbench. Microsoft Kinect Xbox is used as the RGB-D sensor. It is installed around 1 m from the workbench and at 45-degree angle looking down. A Denso robot arm and a Robotiq gripper are used to perform the grapsing.



The MoCap is calibrated follow the official procedure. We define the OptiTrack frame as the \textbf{base frame}, which is set by placing the Ground Plane on the workbench.

For the \textbf{camera-to-base extrinsic calibration,} we use a chess pattern with 4 OptiTrack markers pasted on it. The centroid of the markers intersects the central corner of the pattern. OptiTrack measures the spatial coordinates of the centroid in the base frame and the pixel coordinate of the central corner is computed by recognizing the pattern captured by the Kinect. By moving the pattern around the workspace, multiple coordinate pairs are collected. We solve the perspective-n-point problem to find the required transformation $T_{c2b}$.

For the \textbf{base-to-robot extrinsic calibration}, we have the gripper holding a special calibration device and move through a series of known points in the robot frame. The OptiTrack captures corresponding points in the base frame. Manual rough alignment of the two sets of points by setting the 6 extrinsic parameters is followed by iterative closest point (ICP) to refine the transformation between the two frames. Note that ICP is only used in the calibration, not in the pose estimation.

\begin{figure*}[h]
  \includegraphics[width=\textwidth]{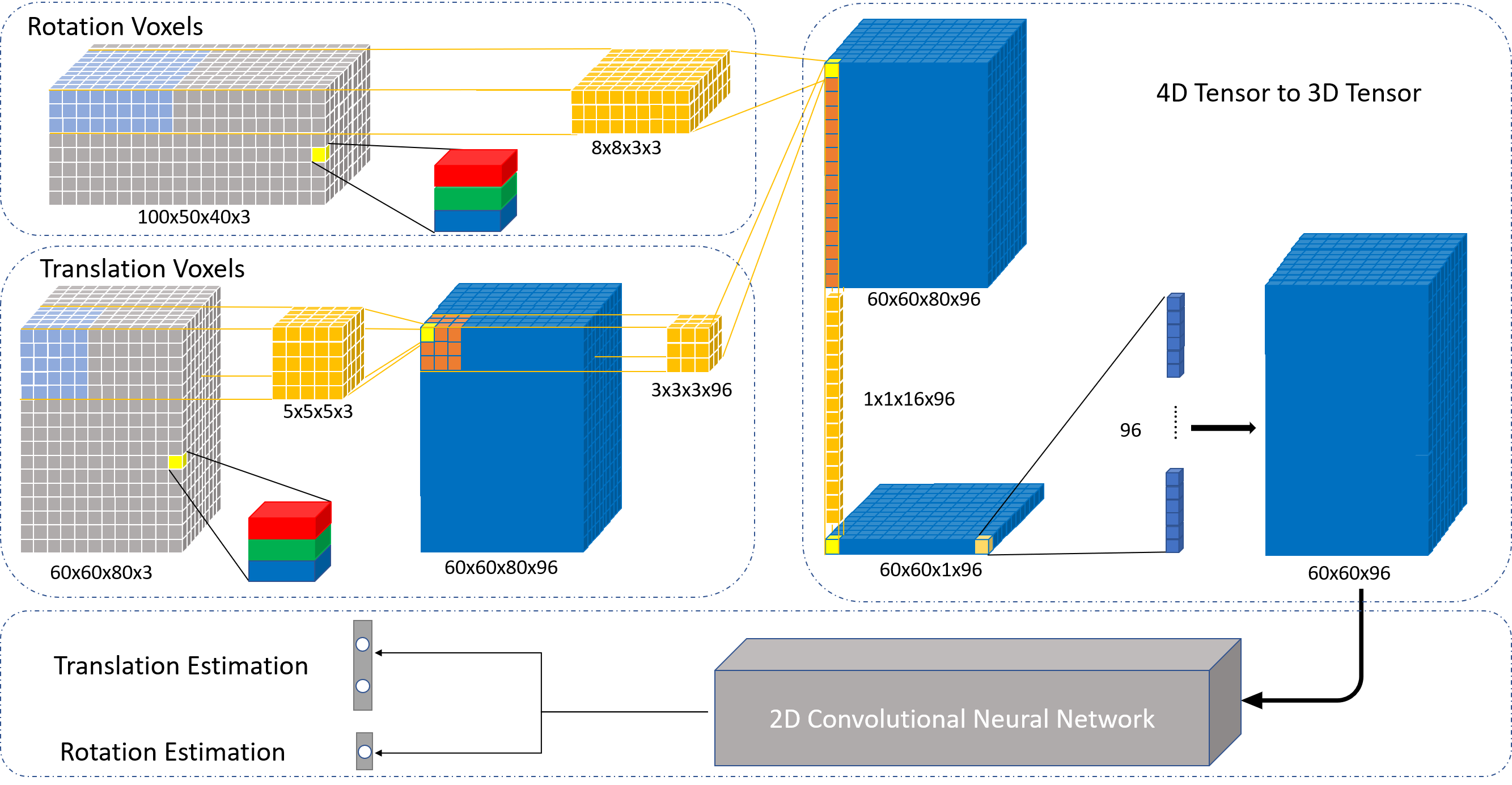}
  \caption{Neural network architecture}
  \label{fig_nn}
\end{figure*}

\section{Data Collection and Automatic Labelling}
\subsection{MoCap for Accurate Pose Detection}
Manual labelling of examples for the training of neural networks is an extremely laborious procedure. Moreover, accurate object pose labels are even more difficult to produce due to errors in measurements.

A common approach towards this is to fix the objects on the workbench and move the cameras around. This has been achieved by using fiducial markers \cite{brachmann_2014_eccv}. However, a common set-up in real-life application often involves a fixed camera and moving targets. 

Some researchers have also used motion capture (MoCap) systems for labelling \cite{wong_2017_iros}, but the objects are still fixed with the cameras moving, which is not essentially different than using the fiducial markers.

We propose to fully utilize a MoCap's capability to provide the pose labels, for it is very accurate (0.2 mm error per marker) and easy to operate with a moving and rotating object. We use 6 OptiTrack 17W cameras to build the MoCap system as shown in Fig. \ref{fig_setup}.

\subsection{Test Objects}
Four objects are selected for training and testing. They are chosen because firstly, they are common domestic objects; secondly, they have irregular geometry, otherwise, pose estimation is less meaningful (a cylinder for example, the geometry is unchanged when rotating around its longitudinal axis); thirdly, they have various shapes of handles which are ideal for grasping.

For each test object, we have an identical pair of them. One is used for data collection and grasp strategy teaching (referred below as "marked object"), the other is used solely for grasping test (referred below as "unmarked object").

OptiTrack markers, which are highly observable to the MoCap cameras, are pasted on the marked object surface. The markers used are much smaller to minimize their effect on the captured colored point clouds. A very interesting observation is that the markers cannot be detected by the Kinect sensor so it appears to be a void in space. The samples we have collected contain many voids amidst colored points due to the limited resolution of the RGB-D point cloud, the effect of markers on the learning is mitigated.

\subsection{Data Collection}
Not the entire 6-DOF space can be explored by a real object because the object always collapses into few stable poses. For example, a mug will not stand on its handle. Therefore, we argue that translation in x and y-axis and rotation about the z-axis are the most important pose parameters and will be the focus of the experiments. We thus simplify the experiment to only consider the situation when the object is standing upright on the workbench. We define the pose with the most prominent geometrical feature pointing to the positive x-axis direction to has 0-degree rotation in yaw. 

To the best of our ability, we repeatedly place the marked object at a random location on the workbench with the most prominent feature pointing at a random direction. As the marked object moves and rotates, synchronized pairs of point clouds captured by the RGB-D sensor and pose labels by the motion capture system are obtained.

In total, we collected around 1500 examples for the purple detergent bottle and around 1000 examples for each of the rest of the objects.

\section{Neural Network Architecture}
\subsection{Architecture}
Instead of one network estimating 6 DOF together, we chose to build a two-stage neural network estimation: the first network takes in a voxelized point cloud with large grid size to estimate the translation of the target object on the workbench. The estimation is then used to crop a subset of the point cloud and a voxelization with finer grid size is performed on the sub-cloud. The new voxelized point cloud is fed into the second stage of the neural networks for orientation estimation. 

The two-stage architecture has three advantages: the estimation of translation and rotation is separated to make trouble-shooting easier; the two stage has to specialize in one task each only; most importantly, the two stage can use voxels with different grid sizes, allowing for a faster translation estimation and a more accurate rotation estimation.

It is possible to replace voxelization with new techniques such as \cite{qi_2018_cvpr} that work on the unordered point clouds directly. However, we prefer more control over the input and to keep the simplicity of the architecture for this is a proof-of-concept effort.

AlexNet is used for 2D images, hence, we topped it with a 3D convolution layer followed by a 3D-2D convolution layer for translation estimation. The 3D-2D convolution layer performs 3D convolution, with the kernel depth dimension matching the input's depth dimension, followed by a dimension squeeze. After this operation, the depth dimension is eliminated, with information in the depth direction encoded. Hence, we can easily apply any 2D CNN backbones to the 3D data after this operation.

Although there are more sophisticated CNNs used in the 2D image detection realm such as VGG and ResNet, we find out that AlexNet \cite{krizhevsky_2012_nips}, which is a relatively simple network, suffices the task requirements with very high accuracy. 

Through experiments, we realized that rotation estimation is a much more challenging task. Hence, the rotation estimation network is similar to the translation estimation counterpart except for an additional 3D convolution layer is added. This significantly improves the performance as more sophisticated feature map is needed for correct interpretation of the orientation of the object. In contrast, the translation estimation where the network only needs to know where the object is most likely to be. 

Instead of predicting the yaw value in degrees or radians, we divide one revolution into 72 equal sectors, each representing 5 degrees. The rotation estimation network estimates in which of the sectors is the object oriented. In practice, 5 degrees error fall well within the tolerance of the gripper. Hence, as long as the rotation estimation network is able to make a correct estimation, it is sufficient for grasping. The starting orientation (yaw = 0 degree) is defined by when the rigid object is first defined in the OptiTrack system. 

\subsection{Loss Functions}
For the translation estimation, the labels are the x, y translation values in meters, with respect to the origin of the base frame. Average L2 Euclidean Loss is used:
$$L_{trans} = \frac{1}{2N}\sum_{i=0}^{N}
(x_{i} - \hat{x}_{i})^{2} + (y_{i} - \hat{y}_{i})^{2}$$
For the rotation estimation, it outputs probability of each of the 72 classes, cross-entropy loss is used to treat it as a classification problem:
$$L_{rotat} = - \frac{1}{N}\sum_{i=0}^{N}\sum_{j=0}^{71}\psi_{ij}\log(\hat{\psi}_{ij})$$
where $\psi_{ij}$ is the probability of the $j^{th}$ class of the $i^{th}$ example, it can either be 0 or 1 and $\sum_{j}\psi_{ij} = 1$. $\hat{\psi}_{ij}$ is the estimated probability of $j^{th}$ class of the $i^{th}$ example. Note that a softmax operation is applied on the output of the fully connected layers of the rotation estimation network to produce $\hat{\psi_{i}}$.

\subsection{Training}
We randomly selected 600 samples from all samples collected for each marked object (except for purple detergent bottle, for which 1200 samples are selected) to avoid any unintentional pattern in the sample collection process with respect to sampling time. Validation and testing set has size 200 samples each. All three sets are exclusive of one another.
Adam Optimizer is used in the training, with $\beta_{1} = 0.9$, $\beta_{2} = 0.999$ and $\epsilon = 10^{-8}$, at a learning rate $\alpha = 10^{-5}$. Both networks are trained from scratch for 100 epochs and on batches of size 50.

\section{Application to Object Grasping}
It is to note that no object model is used at any stage of the pipeline. In fact, no object model is constructed. The only input to the pipeline is the RGB-D point cloud captured by the Kinect sensor, and the only prior required is a set of grasping strategies.

\begin{figure}[htbp]
 \centerline{\includegraphics[width=\columnwidth]{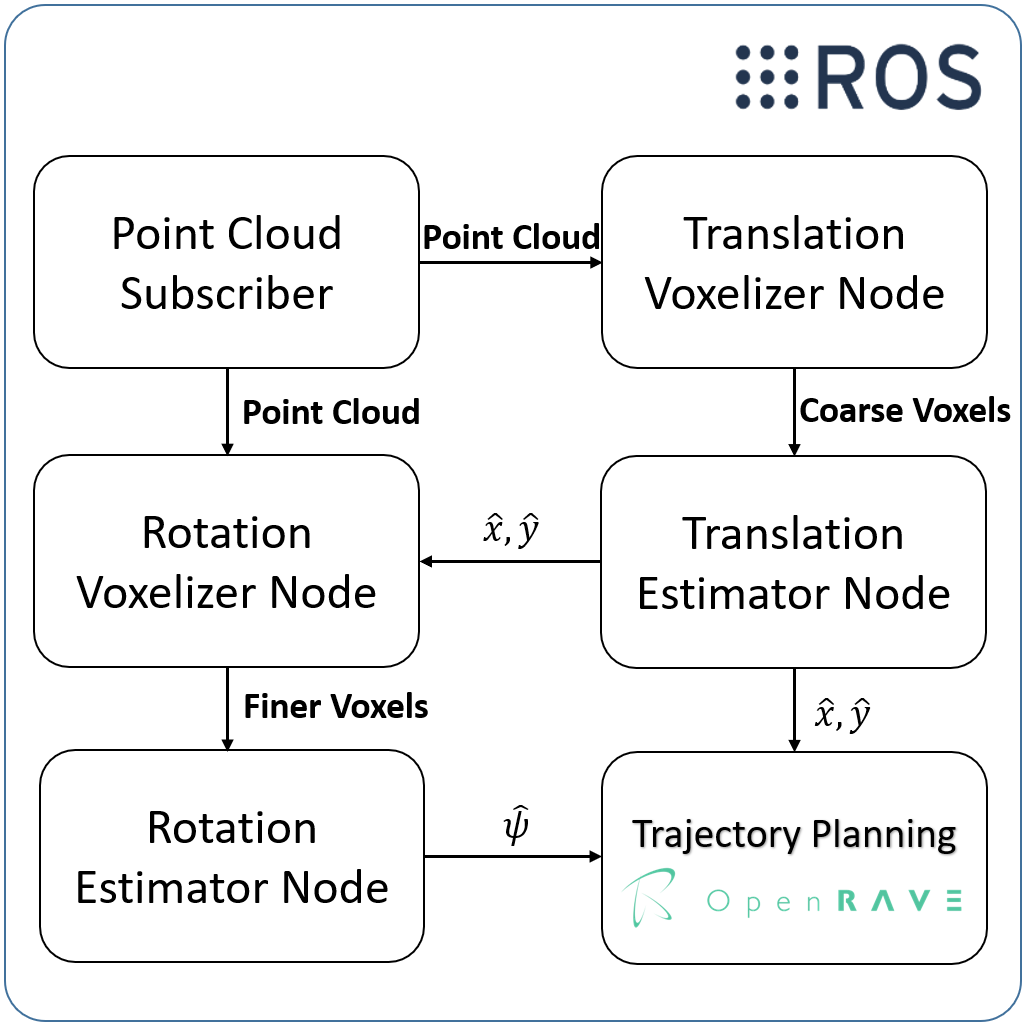}}
 \caption{Pipeline}
 \label{fig_pipeline}
\end{figure}
\subsection{Grasp Strategy Teaching}
We first define a grasp strategy as a gripper-to-object transformations: $T_{g2o}$. $T_{g2o}$ is used to guide the gripper to the grasping pose. One or more grasping strategies are taught for each object for some objects have multiple graspable areas.

For teaching, a marked object is used, where its pose is computed by the MoCap system, from which we obtain $T_{o2b}$ and its inverse $T_{b2o}$. We then guide the gripper to the proper pose and record its pose relative to the robot frame provided by the controller to compute $T_{g2r}$. Therefore, the grasp strategy is obtained:
$$T_{g2o} = T_{g2r}T_{r2b}T_{b2o}$$
where $T_{r2b}$ is the inverse of base-to-robot calibration $T_{b2r}$. 

In practice, we add an intermediate pose in each grasping strategy. The gripper is first moved to the intermediate pose before proceeding to the grasping pose. The procedure to compute the corresponding transformation matrix is the same as that for the grasping pose.

\subsection{Pipeline}
We construct the pipeline (Fig. \ref{fig_pipeline}) to run on ROS. \cite{iai_kinect2} is used to publish the point clouds from the Kinect. The \textbf{point cloud subscriber} takes in the raw point cloud, transform the point cloud into the base frame using the calibrated $T_{c2b}$. Through experiments, we found this transformation aids both estimations. The reason can be that the network has to make estimations that are in the base frame; having the input in the camera frame complicates the problem. 

The \textbf{translation voxelizer node} takes in the transformed point cloud and voxelize a volume of size $100 cm\times50 cm\times40 cm$ above the workbench into the $100\times50\times40$ 3D grids, each grid corresponds to $1 cm^{3}$ space. The volume is defined as the workspace, and its bottom is 5 cm above the workbench surface, as we are not interested in the bottom of the objects, which is often invisible to the RGB-D camera. The dimensions of the 3D voxels are arranged such that its width, height and depth corresponds to x,y and z-axis of the workspace, as if the sensor is placed on the top of the workbench. This bird's eye view is inspired by \cite{chen_2017_cvpr}, which feed networks with projected top view and the front view of the scene. The 45-degree angle placement of the Kinect sensor makes it possible to project the point cloud into both top view and front view. However, it is observed that the top view solely is sufficient for the task.

The \textbf{translation estimator node} then takes in the voxels and publishes the translation estimation after a forward pass. It is important to highlight that the pose estimation of the object is with respect to the base frame instead of the camera frame.

The \textbf{rotation voxelizer node} receives the translation estimations. It crops out a $30 cm\times30 cm\times40 cm$ volume on the transformed point cloud at the location where the object is estimated to be. The volume is voxelized to have a shape of $60\times60\times80$, so each grid represents a $0.125 cm^{3}$ space. For smaller objects such as the black and white mug, the volume can be scaled to be even smaller, for example, $15 cm\times15 cm\times20 cm$. The shape of the voxels is the same regardless of the volume size because the input size to the neural network is fixed. This means a smaller volume will have a finer grid size, improving estimation for small objects.

The finer voxels pass through the rotation estimation network in the \textbf{rotation estimator node}, after which the rotation estimation in the form of the index of the most probable 5-degree sector is published.

In the \textbf{trajectory planning node}, the translation and rotation estimation are combined to compute the transformation matrix of the object frame to the base frame, $T_{o2b}$. The sector index is converted to yaw angle and the parameters of z, roll and pitch are assumed as the only upright objects are considered. Next, we compute the transformation from end effector (gripper) frame to the robot frame $T_{g2r}$:
$$T_{g2r} = T_{b2r}T_{o2b}T_{g2o}$$
where $T_{b2r}$ is obtained from the base-to-robot calibration, $T_{g2o}$ is obtained from grasp strategy teaching. OpenRAVE is used in the grasp node for robot trajectory planning. It takes in the transformation $T_{g2r}$ to compute inverse kinematic (IK) solution for execution. For objects with multiple grasping strategies, strategies without IK solution is disregarded and the closest IK solution from a viable strategy is executed.

\begin{figure}[htbp]
 \centerline{\includegraphics[width=\columnwidth]{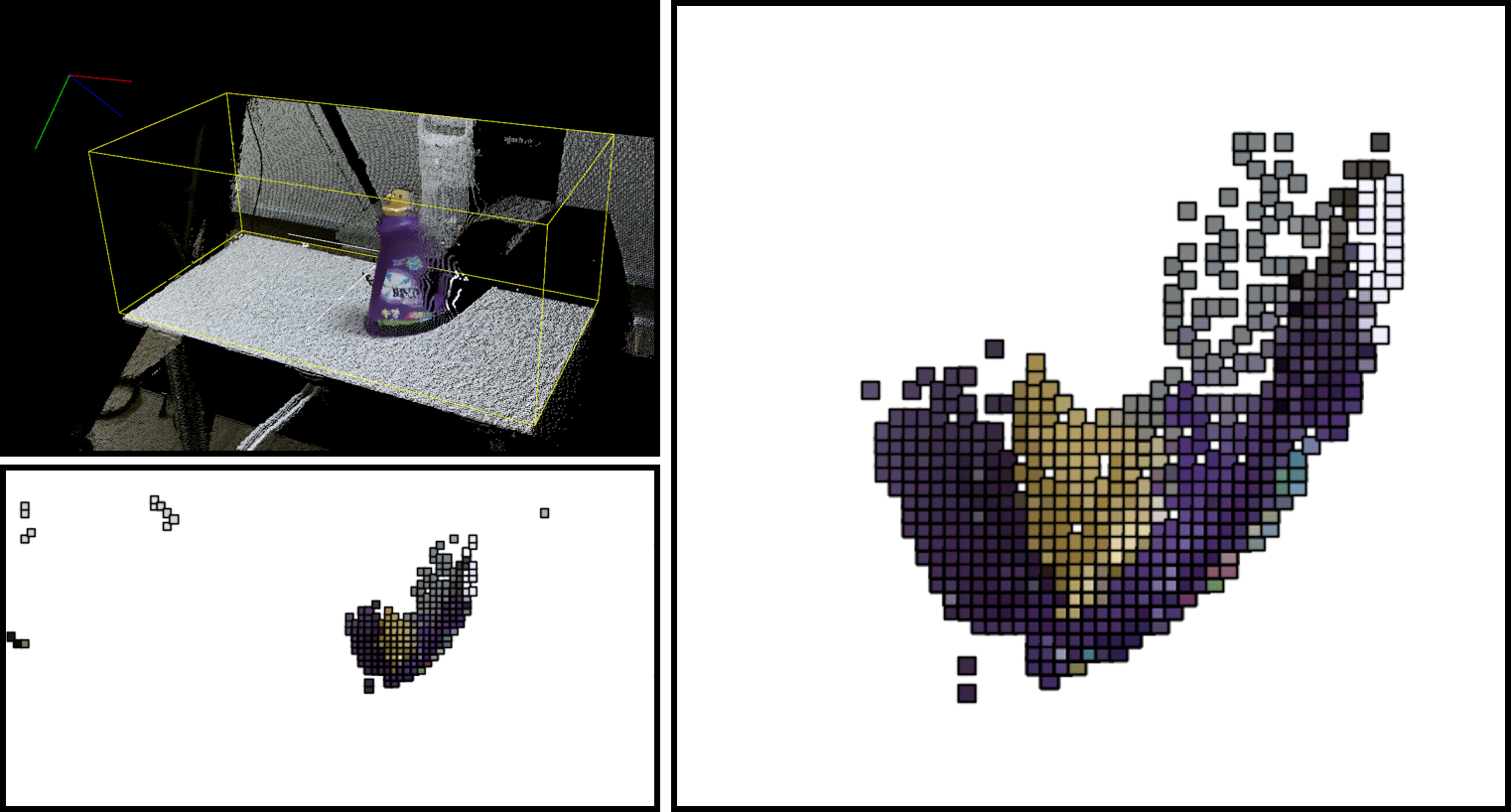}}
 \caption{Raw point cloud (top left), the top view of translation voxels (bottom left) and top view of rotation voxels (right, not the same scale as the translation voxels)}
 \label{fig_voxels}
\end{figure}

\section{Experiment}
\subsection{Testing Set Performance}
After the networks are trained, the testing sets are used to evaluate their performances. For translation estimation, the evaluate metric is the average error among N testing examples:
$$e_{x} = \frac{1}{N}\sum_{i=0}^{N}|(x_{i} - \hat{x}_{i})|,\
  e_{y} = \frac{1}{N}\sum_{i=0}^{N}|(y_{i} - \hat{y}_{i})|$$
where $x_{i}$ and $y_{i}$ and the ground truth labels whereas $\hat{x}_{i}$ and $\hat{y}_{i}$ are the estimations. 

For rotation error, we regard a correct class estimation to have error of 0 degrees. Each class further away from the right class results in additional 5 degrees of error. The average of errors across all testing examples is then computed. To put formally:
$$e_{\psi} = \frac{5}{N}\sum_{i=0}^{N}min(
|\psi_{i} - \hat{\psi_{i}}|, 
|72 - |\psi_{i} - \hat{\psi_{i}}||)$$
where $\psi_{i}$ is the yaw class label and $\hat{\psi_{i}}$ is the estimated yaw class. The class values are from 0 to 71 (total 72 class, each representing a range of 5 degrees). Note that the maximum error possible for one estimation is 180 degrees. 

\begin{figure*}[t]
  \includegraphics[width=\textwidth]{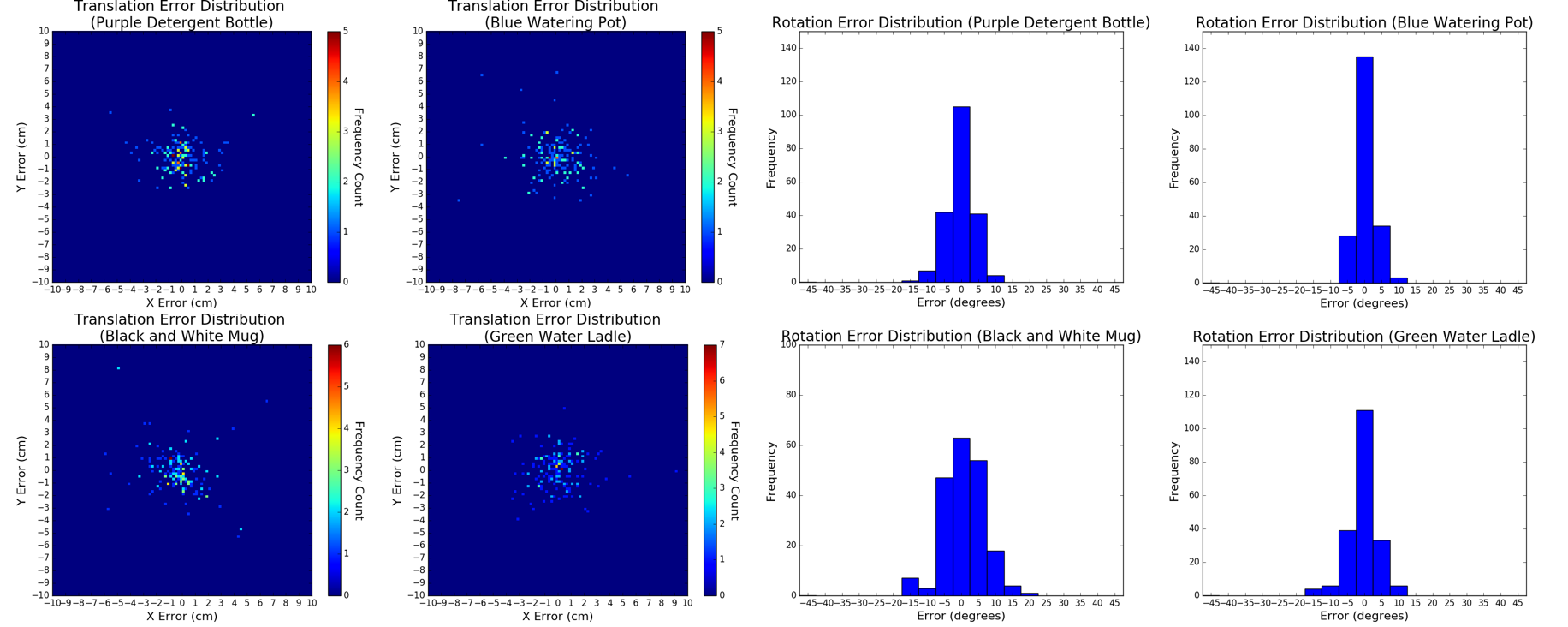}
  \caption{Testing Set Performance}
  \label{fig_test_set_perf}
\end{figure*}
The distribution of testing set errors is visualized in Fig. \ref{fig_test_set_perf}. For translation estimation, most estimation fall within 2 cm error range for both x and y translations. For rotation estimation, the exactly correct class is obtained most of the time; most wrong estimations fall in closest neighboring classes.

\begin{table}[h]
\caption{Testing Set Performance}
\begin{center}
\begin{tabular}{|c|c|c|c|c|}
\hline
\textbf{Object}&\textbf{Training Set Size}&$\mathbf{e_{x}}$\textbf{/cm}&$\mathbf{e_{y}}$\textbf{/cm}&$\mathbf{e_{\psi}}$\textbf{/\textdegree} \\
\hline 
Purple detergent bottle&1200&1.029&0.975&3.43 \\
\hline
Blue watering pot&600&1.140&0.987&1.75 \\ 
\hline
Black and white mug&600&1.190&1.182&5.98 \\
\hline
Green water ladle&600&1.092&1.123&3.35 \\
\hline
\end{tabular}
\label{tab_testing_set_perf}
\end{center}
\end{table}

\subsection{Actual Grasping Performance}
\begin{figure}[htbp]
 \centerline{\includegraphics[width=\columnwidth]{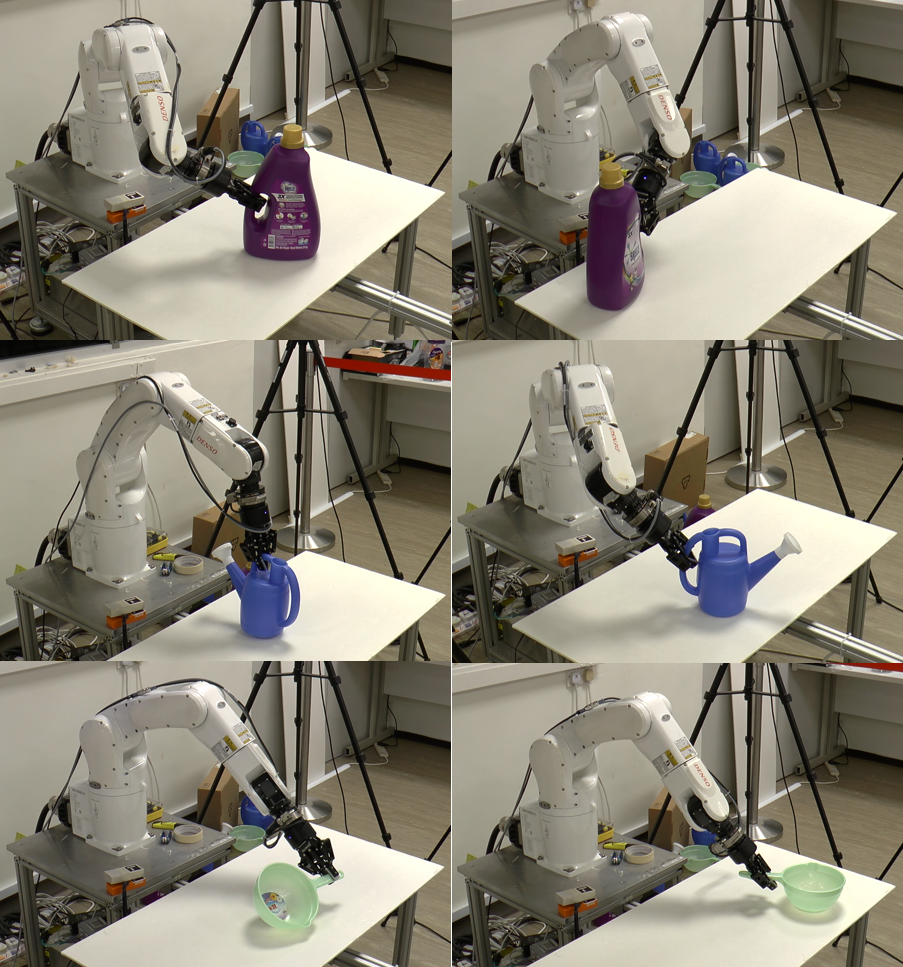}}
 \caption{Pose estimation followed by grapsing of three different objects. A video is available at \url{https://youtu.be/l7QrJYhe2-4}}
 \label{fig_grasp}
\end{figure}
Unlike the samples used in training and testing of the networks, actual grasping is performed on unmarked objects to mimic the real-life applications of object grasping. Also, it is to evaluate the effects of OptiTrack markers on learning of neural networks.

For each unmarked object, we randomly place it on the workbench with its handle pointing at a random direction and let the robot to perform grasping. If the gripper is able to lift the object off the surface of the workbench, the attempt is counted as successful. There are also situations when there is no IK solution found by the trajectory planning due to physical constraints of the robot arm, for example, when the bottle is put at the far corner with its handle point away from the robot, we will not count this as successful but repeat the placement of the object. We collate the first 30 attempts of the robotic arm and record the results.

An important observation from the actual grasping experiment is that the  OptiTrack markers do not affect rotation estimation in most cases. For example, the white sprinkler of the blue watering pot and the handle of the green water ladle are very prominent features. As discussed above, we attribute this to the fact that the markers are invisible to the Kinect sensor.

However, the network tends to learn from OptiTrack markers (which appear voids) where there are very scarce features. For example, we pasted four markers on the opposite side of the handle of the marked purple detergent bottle. When testing against the unmarked bottle, due to the absence of the markers, the network can give totally wrong estimations. 

The black and white mug uses maker base (which are plastic support structures) for easier detection by the MoCap camera, but unlike the markers, the marker base are visible to the Kinect and becomes intruding features, rendering its grasping experiment meaningless and thus not shown. 

\begin{table}[htbp]
\caption{Actual Grapsing Performance}
\begin{center}
\begin{tabular}{|c|c|c|c|}
\hline
\textbf{Object}&\textbf{Attempts}&\textbf{Successes}&\textbf{Success Rate} \\
\hline 
Purple detergent bottle&30&27&90.0\% \\
\hline
Blue watering pot&30&28&93.3\% \\ 
\hline
Green water ladle&30&28&93.3\% \\
\hline
\end{tabular}
\label{tab_actual_grasping_perf}
\end{center}
\end{table}

\section{Conclusion}
We have demonstrated the possibility of building 3D convolution neural networks upon 2D network backbone to directly estimate object's pose. Despite some limitations at the current stage, using deep learning instead of model-based classical approach has shown its potential for highly accurate translation and rotation estimation from a small training set. We have also shown that MoCap can be a new alternative to current pose labelling techniques.

\section{Acknowledgment}
The authors would like to thank Nanyang Technological University for financial support. We are also grateful towards Francisco Suárez-Ruiz, Nicholas Adrian, Hung Pham, Huy Nguyen and Xu Zhang's helpful discussion on the set-up.

\bibliographystyle{IEEEtran}
\bibliography{references}

\end{document}